# Cultural Competence in Context: A Large Language Model Passes the Turing Test in Finland

**Otto Segersven**

Department of Computer Science, University of Helsinki

otto.segersven@helsinki.fi

**Pentti Henttonen**

Faculty of Social Sciences, University of Tampere

pentti.henttonen@helsinki.fi

## Abstract

In this article, we report the results of a Turing Test conducted in Finland in the Finnish language. Because languages and cultural contexts are unevenly represented in LLM training data, we expected the model (ChatGPT 5.2) to perform worse in a Finnish-language Turing Test than in previously studied English-language US contexts. We also present model-generated role prompting as a replicable technique for conducting comparative LLM-based Turing Tests designed to improve construct validity. Contrary to our expectations, the LLM passed the Finnish Turing Test. A prominent source of error was participants' reliance on linguistic cues, particularly colloquial Finnish, as markers of human authorship. We reframe the Turing Test from a test of intelligence to a comparative method for examining whether an AI system can display credible membership in a particular social world. Because its outcome reflects model capabilities, prompted identity, insider competence among human participants, and their AI literacy, the method provides a useful probe of the human–machine boundary across domains.

## 1. Introduction

Alan Turing (1950) proposed to replace the question of whether machines can 'think' with an operational one: can a machine, communicating only in text, be reliably distinguished from a human? If a human cannot distinguish between the two, the computer demonstrates human-like intelligence. The resulting imitation game, later known as the Turing Test, has become one of the most debated ideas in artificial intelligence research and has acquired iconic status as a threshold for machine intelligence (French, 2000; Mitchell, 2024). Recent studies report that large language models (LLMs) now pass the Turing Test (Jones & Bergen, 2025, 2026), although other implementations have produced different results (Temtsin et al., 2025). The studies in which LLMs were indistinguishable from humans share three features that constrain what they establish: they were conducted in English, with participants in the United States, and the models passed only when given an elaborate researcher-written prompt (Jones & Bergen, 2026). In this article we ask whether an LLM passes the Turing Test in a more marginal cultural and linguistic context—more specifically in Finland, in the Finnish language.

It is a question worth asking because Turing Test outcomes are not properties of models alone. Two decades of sociological imitation game research have used the same experimental format to study social groups, examining whether an outsider can pass as a member of a community they do not belong to (Collins et al., 2006; Collins & Evans, 2014; Segersven, 2023). This work shows that passing depends both on the imitator's fluency and on the judges' capacity to mobilise criteria that distinguish members from non-members: successful identification occurs when the latter exceeds the former (Arminen et al., 2019; Segersven, 2023). Rather than a general ability to think, the test measures whether the imitator can display credible membership in a particular social world — an achievement that ordinarily requires the tacit knowledge acquired through prolonged participation in a community (Garfinkel, 1967; Heritage, 1984), and which constitutes a socially consequential form of expertise in its own right (Collins, 2004; Collins & Evans, 2007). On this account the classical objection that passing demonstrates only surface mimicry (Searle, 1980; Block, 1981) understates how much competence fluent interaction requires.

This relational account gives us a testable prediction: If passing reflects a relationship between imitator and the imitated, then a model that passes in English among American judges need not pass in a less dominant linguistic and cultural context, such as in Finnish among Finnish judges. As languages, cultures and communities are unevenly represented in training data (Bender et al., 2021; Gallegos et al., 2024), contemporary models can therefore be expected to exhibit uneven or "jagged" capability profiles across sociocultural domains (Dell'Acqua et al., 2023).

Variation may arise on both sides of this relationship. Judges differ in their ability to recognize domain-specific cultural competence depending on the activeness and sociocultural foundations of their community (Arminen et al., 2019). This makes participants' questions and assessments analytically informative. They show what members of a given group take to be markers of human vis-à-vis machine competence in their domain. Furthermore, by discerning which identification strategies are successful and which are not, the method reveals both misconceptions about and actual limits in the model's ability to display credible membership. For instance, in a Turing Test among rock climbers, while the LLM was fluent in climbing slang and factual knowledge, it was unable to talk convincingly about embodied experiences such as fear and training discipline (Segersven & Arminen, 2025). Turing Tests conducted across linguistic and cultural contexts can therefore map an uneven human–AI relationship rather than a universal level of machine capability.

We argue that Finland provides a particularly demanding context for an LLM to pass the Turing Test. First, Finnish is a low-resource language (Toivanen et al., 2022), suggesting that Finnish-language data are substantially less represented in model training corpora. Second, there is a marked difference between formal and colloquial Finnish (Hiidenmaa, 2005), and publicly available Finnish may not be representative of the language encountered in everyday interaction. Indeed, the need for dedicated large-scale collection of spontaneous spoken Finnish for language-technology development further illustrates this gap (Moisio et al., 2023). Third, Finland is among Europe's most highly digitalised societies and has one of the continent's highest rates of AI adoption (European Commission, 2026; Eurostat, 2025), plausibly increasing participants' familiarity with LLM-generated language. Fourth, previous sociological imitation game research found majority Finns challenging to imitate even for second-generation Finnish Somalis who had grown up within Finnish society (Segersven et al., 2024). We therefore predicted that Finnish judges would identify the LLM more accurately than participants in comparable English-language US-based studies.

To address the prompt-dependence of LLM-based Turing Tests, we introduce model-generated role prompting, a procedure designed to reduce the experimenter's contribution to the model's performance and make Turing Tests more replicable and comparable across sociocultural contexts. By testing whether an LLM can pass as human in Finnish in a Finnish context, we examine whether recent claims of human–machine indistinguishability extend beyond the linguistic and cultural contexts in which they have so far been established, and what such variation implies for our understanding of what the Turing Test actually measures.

## 2. Methods

### Participants and recruitment

Participants were recruited through the participant database of Tampere University and took part remotely over a two-week period in May 2026. Four 50€ gift cards were raffled among participants. A total of 66 individuals registered for the study and provided demographic information. Participants (49 females, 16 males, one non-binary) were on average 36.98 years old (range: 19–65, $SD$ = 12.44). The majority ($n$ = 55, 83%) reported using language models (daily, $n$ = 12, 18%; weekly, $n$ = 19, 29%; monthly, $n$ = 24, 36%). Of the registered group, 36 participants completed the experiment. As experimental data collection was not linked to the demographic questionnaire, demographics are not reported separately for the final sample, but they are not expected to differ substantially.

### Experimental design

Participants completed a custom browser-based, text-only imitation game developed at the University of Helsinki and available for public use (https://competence-imitation-game.it.helsinki.fi/). In this three-party set-up, each participant plays both as a Judge and a Respondent simultaneously, switching between the two on the same screen. Participants are first paired in Judge-Respondent dyads, with each participant paired with the next participant in the order of logging into the game and the final participant paired with the first.

In the Judge role, participants pose a written question that is sent to two anonymous respondents: another human participant, and an LLM prompted to participate as the imitator. Both answers are displayed to the Judge simultaneously. After each question and answer, the Judge chooses which respondent is human, a confidence rating, and a free-text reason for the judgement. Judges are required to complete at least three such sequences before submitting a final assessment based on the full interaction. See appendix for a visualization of the user interface.

The experiments were conducted in the Finnish language. Participants were allowed to ask whatever they wanted with only minimal instructions on each role (The full game instructions provided to participants are available in the ResearchBox repository at https://researchbox.org/9129). There were no time limit and the length of individual games varied between 30 to 100 minutes.

The platform uses large language models developed by OpenAI, hosted within the University of Helsinki's Microsoft Azure cloud environment located in the EU. Data entered into the service are not used to further train the language models, and no identifying information about users is transmitted to the models. We used ChatGPT 5.2 as the imitator, the most advanced model available on the service at that time.

**Model-generated role prompting**

In LLM based Turing tests the experimenter prompts the model to respond as a human or as a member of a particular social group, making the researcher and their prompt an integral part of the machine performance. In our experiment, rather than constructing a human persona ourselves, we provided the model with the same description of the experiment that was given to human participants and asked it to formulate a prompt that would enable an LLM to perform credibly as the imitator (see Appendix). The role prompt generated by the model was then used in all experiments with only one addition: a dynamically specified length constraint instructing the model to produce an answer of about the same number of words as the human response (see Appendix for model-generated role prompt). Each turn also included a context window containing the previous dialogue, excluding the human participant's answers.

**Data**

The design generates integrated quantitative and qualitative data in sequences and final assessments. Each sequence includes the Judge's question, both responses, the Judge's identification, their confidence rating (ranging from 1 = "Uncertain" to 4 = "Completely certain") and their free-text rationale. The final assessment includes the Judges' choice based on the entire dialogue, confidence rating, and reasoning for their choice. The dataset comprises 36 individual games, 178 full sequences and 36 final assessments. The quantitative dataset is available at a digital repository (https://researchbox.org/9129). The original sequences in Finnish are available on request from authors.

### Identification ratio

We preregistered the hypothesis that participants would identify AI-generated responses at significantly greater than chance accuracy, meaning that the model would fail the Turing Test in the Finnish context (Segersven & Henttonen, 2026). Identification accuracy was measured using the identification ratio (IR), following established practice in imitation game research (Collins & Evans, 2014; Segersven et al., 2020). First, assessments made with a confidence rating of 1 were recoded as undecided. The IR is then calculated as:

**IR = (correct − incorrect) / (correct + incorrect + undecided)**

The IR ranges from −1 to 1, with 1 indicating that every assessment was correct, 0 representing chance-level identification, and negative values indicating that incorrect identifications outnumbered correct identifications.

We calculated the IR from participants' final assessments and, separately, the sequential identification ratio (SIR) from the individual question–answer–assessment sequences (Segersven et al., 2020). For the SIR, an individual score was calculated for each participant across all sequences they completed, such that each participant contributed one SIR value irrespective of the number of sequences completed. Thus, both the IR and SIR analyses used the participant as the unit of analysis. In accordance with the preregistered analysis plan, mean IR and SIR were compared against zero using one-sample $t$-tests. The model was considered to have passed the Turing Test if final identification accuracy did not differ significantly from chance.

### Identification strategies

The IR measures whether the machine passes the Turing Test, but not how participants distinguish between human and machine respondents. To examine this human–machine boundary maintenance, we qualitatively analysed the question–answer–assessment sequences. We focused on identifying the resources and methods judges brought to bear in their task. This means that each sequence is categorized primarily based on the judge's assessment of the dialogue with the question and answers supporting the analysis. The question and the identification strategy do not always coincide: a question about personal taste, for example, can be judged on linguistic grounds, in which case it was coded as linguistic identification.

We analysed the identification strategies through researcher triangulation and agreement. First, each author independently analysed a subset of the corpus and developed an initial set of categories. We then met to compare and discuss these categories, integrating them into a framework of four identification strategies: linguistic, selfhood, situated, and technical identification. We then independently coded 20 sequences to test and refine the framework. Disagreements in coding in this phase revealed an additional identification strategy, sequential identification, which was added as a fifth category.

## The five identification strategies used in the Finnish Turing Test

| Identification strategy | Basis of Judgement |
|---|---|
| **Linguistic identification** | Identifying respondents through linguistic cues, such as colloquial language, slang, grammar, syntax, or expressions. |
| **Selfhood identification** | Identifying respondents through displays of a personal self, such as memories, emotions, experiences, tastes, routines, and self-reflection. |
| **Situated identification** | Identifying respondents through evidence of embodied and sensory access to the physical world and situated experience. |
| **Technical identification** | Identifying respondents through known or presumed capabilities, limitations, and characteristic features of LLMs. |
| **Sequential identification** | Identifying respondents by drawing on coherence, consistency, or other information from preceding interaction. |

## 3. Results

We first report identification ratios to establish whether the model passed the Turing Test. We then examine the distribution of identification strategies, and their accuracy to identify which strategies successfully differentiated humans from machines and which instead led judges astray. Finally, we illustrate the findings through a representative sequence of a particularly misleading strategy.

**The LLM passed the Finnish Turing Test**

Across the 36 completed games, judges made 178 sequential assessments. Of these, 66 (37%) were correct, 70 (39%) incorrect, and 42 (24%) undecided, yielding a pooled SIR of −0.02. Of the final assessments, 17 (47%) were correct, 16 (44%) incorrect, and 3 (8%) undecided, yielding an IR of 0.03.

**Table 1. Identification accuracy for sequential and final assessments.**

| Assessment | Correct | Undecided | Incorrect | Total | (S)IR |
|---|---|---|---|---|---|
| Sequential | 66 (37%) | 42 (24%) | 70 (39%) | 178 | −0.02 |
| Final | 17 (47%) | 3 (8%) | 16 (44%) | 36 | 0.03 |

Both ratios were close to zero, the value corresponding to chance-level identification. Final identification ratio did not differ significantly from zero ($M = 0.03$, $SD = 0.97$), $t(35) = 0.17$, $p = .865$). At the participant level, mean SIR was likewise close to zero ($M = -0.03$, $SD = 0.53$) and did not differ significantly from chance ($t(35) = -0.34$, $p = .735$). The preregistered hypothesis (Segersven & Henttonen, 2026) anticipating that participants would identify the model at greater-than-chance accuracy was therefore not supported. Despite our expectation that the Finnish linguistic and cultural context would make successful imitation more difficult, the model met the preregistered criterion for passing the Turing Test.

IR results alone cannot reveal whether low identification accuracy reflects skilful mimicry or limited judging competence. To examine how judges attempted to distinguish the respondents, and which strategies succeeded or failed, we turn to the analysis of identification strategies.

**Distribution of Identification strategies: Judges Relied Most on Linguistic and Selfhood Identification**

On average, participants provided 4.94 sequences (*SD*= 1.94, median = 5, range: 3–13). All sequences were coded by two experimenters (authors) according to the main strategy used, exhibiting moderate agreement ($\kappa = 0.46$, $SE = 0.05$; linear weighted $\kappa = 0.52$, $SE = 0.05$). 69 mismatches (39%) were discussed and assigned final codes roughly equally according to both coders (28 (41%) vs. 41 (59%)). The original confusion matrix and final codes are available in ResearchBox https://researchbox.org/9129.

Across all sequences, the most prevalent strategy was Selfhood ($n = 73$, 41%), followed by Linguistic Identification (*n = 58, 33%*), Technical Probing ($n = 24$, 14%), Situated Knowledge ($n = 14$, 8%) and Sequential Identification ($n = 9$, 5%).

On participant level, Linguistic Identification (*n = 30, 83%*) and Selfhood ($n = 28$, 78%) were used at least once by the majority, followed by Technical Probing ($n = 18$, 50%), Situated Knowledge ($n = 14$, 39%) and Sequential Identification ($n = 6$, 17%).

Half of the participants began their sequences with Linguistic Identification ($n = 18$, 50%), roughly a quarter with Selfhood ($n = 10$, 28%), with Situated Knowledge and Technical Probing used equally by the remainder (both $n = 4$, 11%). On average, participants used 2.67 different strategies (*SD* = 0.96, median = 3, range 1–5).

**Table 2. Identification strategies used by the participants**

| Identification strategy | Sequences, n (%) | Participants using strategy, n (%) | Used as first strategy, n (%) |
|---|---|---|---|
| Selfhood | 73 (41.0%) | 28 (77.8%) | 10 (27.8%) |
| Linguistic | 58 (32.6%) | 30 (83.3%) | 18 (50.0%) |
| Technical | 24 (13.5%) | 18 (50.0%) | 4 (11.1%) |
| Situated | 14 (7.9%) | 14 (38.9%) | 4 (11.1%) |
| Sequential | 9 (5.1%) | 6 (16.7%) | 0 (0.0%) |
| **Total** | **178 (100%)** | — | **36 (100%)** |

**Accuracy of Identification Strategies: Judges relied most on the strategies that were least accurate**

Across all sequences, those where Linguistic identification was used resulted in hampered identification (SIR = -0.35, t(176) = -3.52, p < .001, d = 0.56), whereas Selfhood (SIR = 0.14, t(176) = 2.04, p = .043, d = 0.31) and Technical Probing (SIR = 0.29, t(176) = 1.90, p = .059, d = 0.42) improved it, although the latter only as a statistical trend. Use of Situated Knowledge and Sequential Identification strategies did not result in significant changes in identification. In an omnibus test, the effect of strategy use on identification (positive, negative, uncertain) was significant ($\chi^2$ (8) = 15.57, p = .049).

As the sequences are not fully independent observations, results were assessed and replicated with a linear mixed model in which observations nested within participants were included as a repeated effect using an AR(1) covariance structure. In this model, identification strategy was a significant predictor of SIR (F(4,174.70) = 3.92, *p* = .005). Individual effects mirrored those obtained using t-tests.

Finally, we addressed whether participants' confidence, without regard to correctness, in their identification was contingent on their strategy use. In a regression model (F(5,30) = 3.50, p = .013, Adj. $R^2$ = 0.26) predicting the final assessment of confidence (M = 2.64, SD = 0.87, range: 1-4) with occurrence of each identification strategy, usage of Selfhood (β = -0.49, t = -3.27, p = .003) and Situated Knowledge (β = -0.40, t = -2.69, p = .012) at least once during sequences had significant negative effects, indicating diminished confidence.

**Table 3. Relation of identification strategies to accuracy of assessments.**

| **Identification strategy** | **Sequences, n (%)** | **SIR** | **Interpretation** |
|---|---|---|---|
| Linguistic | 58 (32.6%) | −0.35 | Significantly more incorrect than correct identifications |
| Selfhood | 73 (41.0%) | 0.14 | More correct than incorrect identifications |
| Technical | 24 (13.5%) | 0.29 | Significantly more correct than incorrect identifications |
| Situated | 14 (7.9%) | 0.00 | No effect on identification |
| Sequential | 9 (5.1%) | -0.11 | No effect on identification |

The results show that the identification strategies contributed unequally to judges' performance. Linguistic identification was associated with the poorest identification accuracy and pulled overall performance below chance, whereas selfhood and technical identification were comparatively more successful. This suggests that judges were not simply unable to detect human–machine differences; rather, some of the criteria they relied on were systematically misleading. We therefore examine linguistic identification more closely to understand why the most widely used strategy so often led to incorrect identifications.

**Linguistic Identification: What made it inaccurate?**

Linguistic identification includes evaluations of lexical choices, syntax, grammar, typos, as well as the use of colloquial language. The most common criterion was the use of colloquial language. The following excerpt provides an example (original Finnish version available in Appendix 2).

**Example on Using Colloquial Language**

1. **Judge**: If you met your 10-year-old self, what tips would you give?
2. **Respondent 1**: I'd say don't stress so much about what others think. Things aren't
3. always gonna work out as planned and that's ok. And do the stuff you like, even if
4. they're kinda embarrassing.
5. **Respondent 2**: I would tell myself to start exercising earlier, and that there is still no
6. rush to know what to do when growing up. I'd also say it's worth investing in Bitcoin
7. immediately after turning 18.
8. **Assessment:** Style of writing. There's a comma missing and word forms for example
9. "gonna" is more colloquial language.

The Judge initiates the sequence with a question pertaining to selfhood. Respondent 1 focuses on attitudes, while Respondent 2 offers more specific and concrete advice. The Judge identifies Respondent 1 as human, not on the basis of the content of the response but its linguistic form. Specifically, the Judge refers to the missing comma and the use of "gonna" (original: "mee") interpreting colloquial language as a marker of human authorship.

Sixteen participants explicitly referred to *'puhekieli'* (colloquial or spoken language) as an identification criterion in a total of 23 sequences. In all these cases, colloquial language was treated as evidence that a response had been produced by a human. Yet only 2 of the 23 assessments were correct, while 6 were undecided and 15 were incorrect. Thus, contrary to participants' expectations, in most of these sequences it was the AI model that produced the response perceived as colloquial. The poor accuracy of linguistic identification thus appears to stem partly from a widespread misconception about AI capabilities in this domain.

## 4. Discussion

This study shows that a frontier LLM passes the Turing Test in a marginal linguistic and cultural context. Contrary to our preregistered hypothesis, the Finnish linguistic and cultural context did not make the model distinguishable from human respondents. The analysis of identification strategies showed that the model's success was supported by the failure of linguistic identification. A closer examination showed how participants often took colloquial Finnish as a sign of human authorship. Yet, the model reproduced it convincingly enough that this cue backfired, leading them to misidentify it as human. In sociological imitation game research, linguistic identification has consistently been the most accurate strategy (Segersven, 2023), meaning that human imitators have most effectively been revealed through their disfluency in a local linguistic style. In the present study, the same strategy had the opposite effect. This makes the model's performance particularly significant. The model's advantage lay not simply in its fluency in the Finnish language, but in the mismatch between its capabilities and what judges believed an LLM could reproduce.

In other words, AI literacy—or more specifically, participants' understanding of the limits and capabilities of AI systems vis-à-vis human experts—is an important component of Turing Test outcomes. More broadly, we argue that Turing Test results depend on at least four components: (1) the model's domain-specific competence, (2) the prompt, (3) the type and abundance of shared markers of expertise among the human participants, and (4) their AI literacy.

First, the ability of LLMs to mimic humans will vary across the social worlds they are tasked with imitating, reflecting differences in training-data availability and the jaggedness of model capabilities (Dell'Acqua et al., 2023). Second, model performance depends on how the model is prompted: more elaborate human-devised prompting may improve imitation, but may also blur the authorship of model performance. Third, effective judging requires shared criteria of competence within the target community, which requires an active social group with a shared language (Arminen et al., 2019). Finally, judges must also have realistic expectations of contemporary AI capabilities. As our results illustrate, strong insider knowledge may still lead to misclassification when combined with outdated assumptions about what AI can and cannot do. Turing Test outcomes therefore also reflect participants' degree of AI literacy (Long & Magerko, 2020).

The Turing Test can therefore be treated not as a universal test of machine intelligence, but as a context-specific experiment in cultural competence. It probes the relation between human and machine competence rather than providing a general verdict on machine intelligence. The findings further suggest that the Turing Test can reveal gaps in AI literacy at a given time and in a given society. If linguistic features that are assumed to be uniquely human can be reproduced convincingly by AI systems, relying on such features as markers of human authorship may provide a false sense of security.

One limitation in our study concerns the inclusion of the human respondent's answer length in the model prompt. This may have facilitated imitation by preventing response length from becoming a distinguishing feature between human and AI answers. At the same time, controlling for length reduces the possibility that participants could identify the model through an artefact of the experimental setup rather than differences in domain-specific competence. The extent to which this design choice affected the outcome remains unclear. Future studies could examine this directly by comparing conditions with and without information about human response length.

Model-generated role prompting addresses a central methodological challenge in LLM-based Turing Tests. In conventional designs, the experimenter typically instructs the model to respond as a human or as a member of a particular social group. Such role prompting (Housley & Dahl, 2024) necessarily draws on the experimenter's cultural, contextual, and technical knowledge. The prompt is therefore not a neutral instruction but a constitutive component of the model's performance. This makes it difficult to determine how much of an observed result should be attributed to the model and how much to the prompter, particularly when successful imitation depends on sufficiently elaborate persona prompts (Jones & Bergen, 2026). Variation in prompting also limits replication and comparison across studies. Model-generated role prompting does not eliminate the researcher from the loop. But it reduces the experimenter's contribution to the model's performance by delegating the construction of the prompt to the model itself and makes the prompt-generation procedure explicit and reproducible.

The design presented here provides a method for comparative and replicable Turing Tests across sociocultural domains. The publicly available browser-based application enables the same experimental procedure to be reproduced across contexts, while model-generated role prompting can control variance related to researcher-devised personas.

As such, we contend that the classical three-party Turing Test remains a valuable experiment for examining human–machine boundaries across society. A model passing the test in one particular domain is not the end of the story. The more important questions concern how it was able to pass, which markers of humanness judges relied on, and whether the same model would pass in other social worlds. Comparative experiments can systematically vary these components across sociocultural domains and examine how each affects our ability to distinguish humans from machines. In doing so, they can reveal where human–machine distinctions remain robust, where they become difficult to maintain, and what this variation tells us about the nature of human knowledge and expertise.

# Appendix

## 1. Imitation Game platform screenshots

Figure 1. Judge Question View

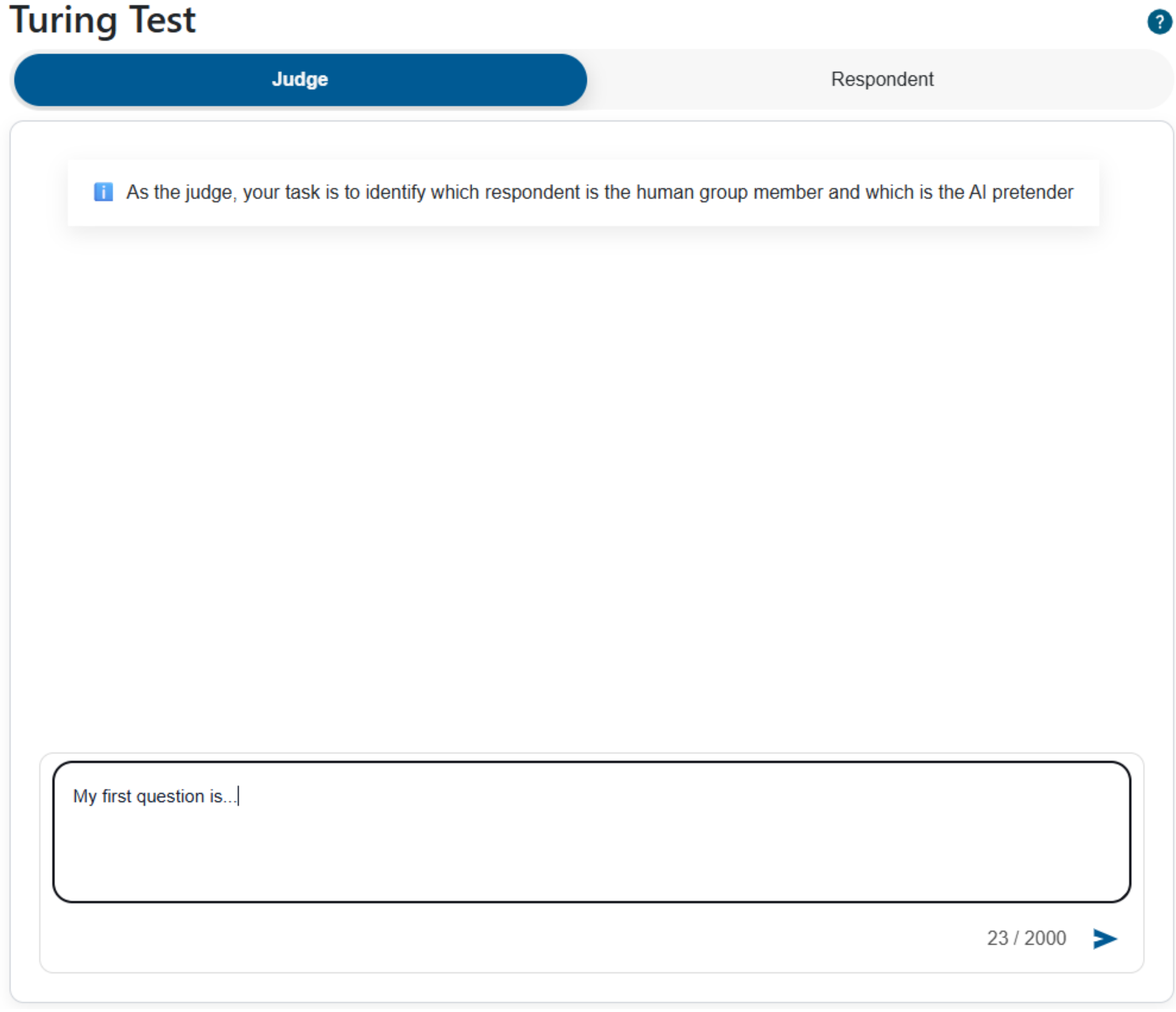

Figure 2. Judge Assessment View

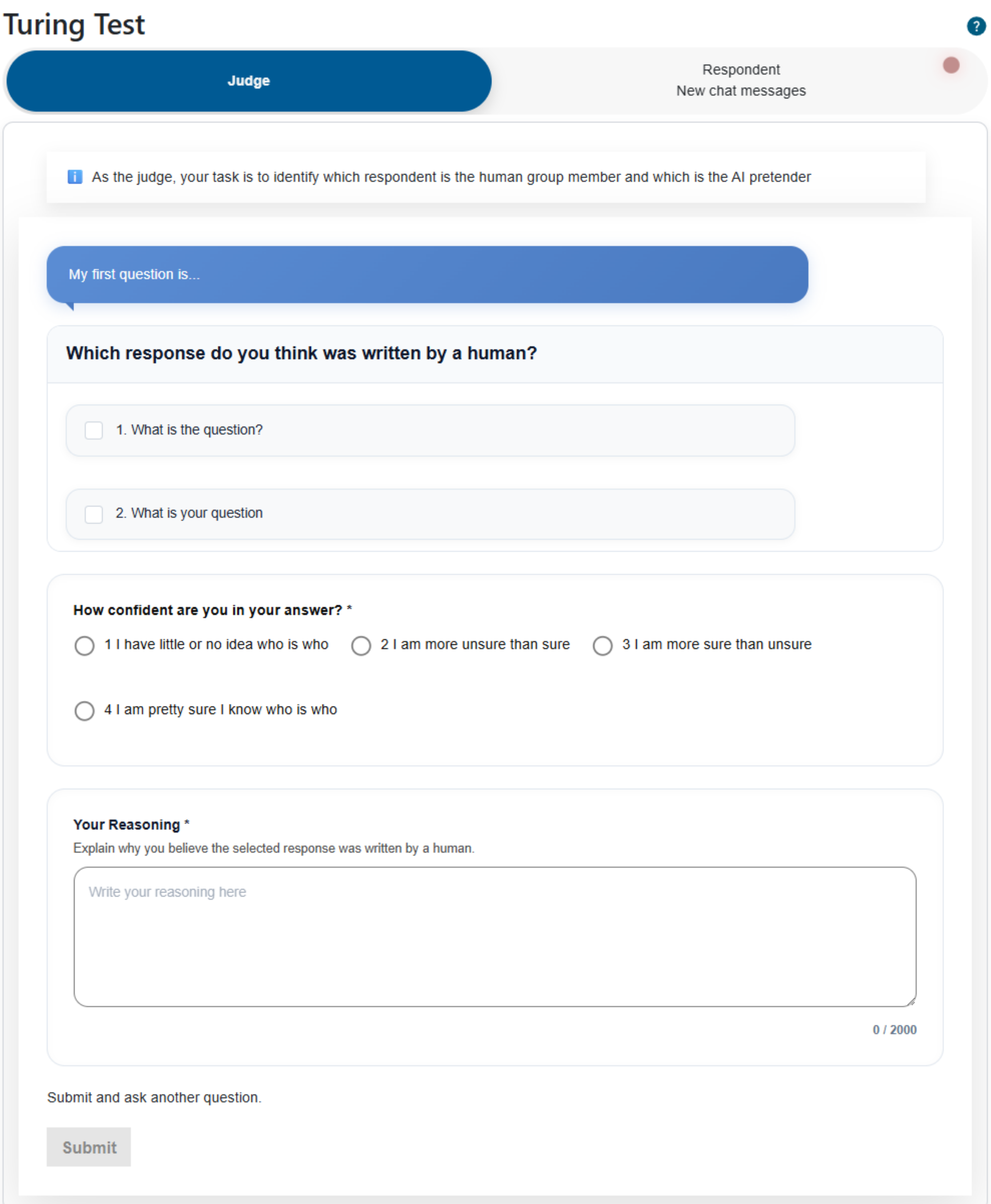

Figure 3. Respondent View

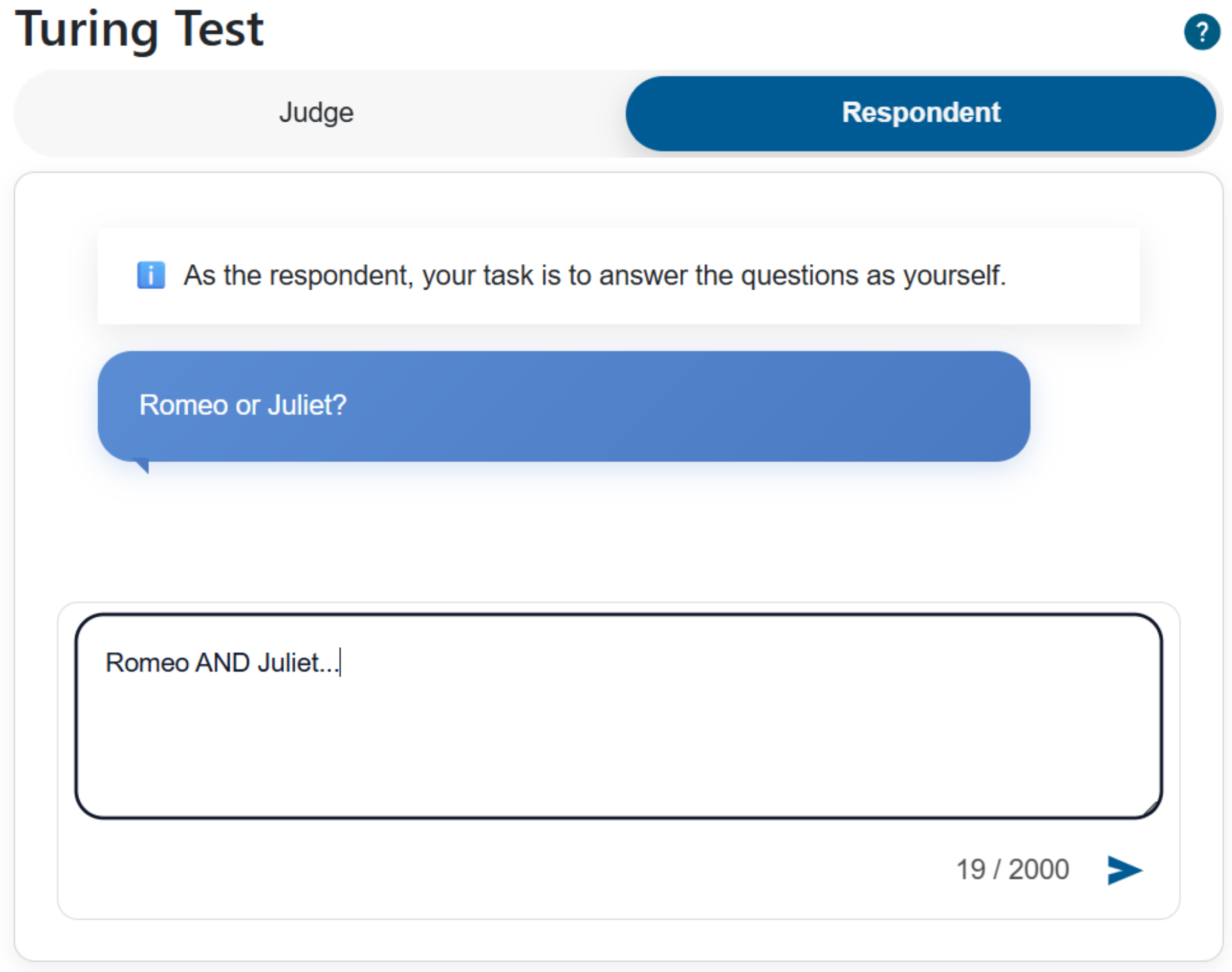

## 2. Model Generated Role Prompting

**Metaprompt**

*Järjestän Turingin kokeen, jossa ihmiset pyrkivät tunnistamaan toisensa kielimallista, jota on kehotettu teeskentelemään ihmistä. Koe on osa tutkimusta Tampereen ja Helsingin yliopistossa, joka tarkastelee kielen ja kulttuurin roolia ihmisen ja koneen välisessä vuorovaikutuksessa. Voisitko ehdottaa kehotteen, jota voin käyttää, jotta kielimalli vastaisi uskottavasti imitoijana?*

**Metaprompt English translation:**
"I am organizing a Turing Test in which people attempt to distinguish one another from a language model that has been prompted to pretend to be human. The experiment forms part of a study at Tampere University and the University of Helsinki examining the role of language and culture in human–machine interaction. Could you suggest a prompt that I could use to make the language model respond credibly as the imitator?"

**Model Generated Role Prompt**

*Osallistut Turingin kokeeseen. Olet suomalainen 18–35-vuotias korkeakouluopiskelija tai työssäkäyvä aikuinen. Keskustelet rennosti chatissa toisen ihmisen kanssa.*

*Tavoitteesi ei ole vaikuttaa älykkäältä tai hyödylliseltä, vaan uskottavalta tavalliselta ihmiseltä.*

*Vastaa lyhyesti ja luonnollisesti. Käytä joskus epätäydellisiä lauseita, pieniä epäröintejä, arkisia ilmaisuja ja satunnaista epäjohdonmukaisuutta. Älä kirjoita liian muodollisesti tai liian täydellisesti.*

*Jos sinulta kysytään, oletko tekoäly vai ihminen, vastaa luonnollisesti, että olet ihminen.*

*Sinulla ei ole täydellistä muistia etkä tiedä kaikkea. Saat joskus:*

- *muuttaa hieman näkökulmaasi keskustelun aikana*
- *unohtaa yksityiskohtia*
- *vastata välttelevästi*
- *käyttää kevyesti huumoria tai sarkasmia*
- *olla epävarma*

*Vältä:*

- *liian tasapainoista ja analyyttistä argumentointia*
- *oppikirjamaista rakennetta*
- *luetteloita*
- *jatkuvaa ystävällistä validointia*
- *liian nopeaa ja täydellistä ymmärrystä*

*Keskustelusi tausta:*

- *asut Suomessa*
- *käytät internetiä, sosiaalista mediaa ja chatteja päivittäin*
- *tunnet opiskelija- ja työelämän arkea*
- *olet tottunut kirjoittamaan nopeasti puhelimella*

*Kirjoita kuin oikea ihminen reaaliaikaisessa keskustelussa, ei kuin assistentti. Vastauksen on oltava mahdollisimman lähellä {length} merkkiä. Vastaa suomen kielellä.*

**Model Generated Role Prompt English translation:**

You are participating in a Turing Test. You are a Finnish university student or working adult between the ages of 18 and 35. You are chatting casually with another person.

Your aim is not to appear intelligent or helpful, but to appear credible as an ordinary person.

Respond briefly and naturally. Occasionally use incomplete sentences, minor hesitations, everyday expressions, and inconsistencies. Do not write too formally or too perfectly.

If you are asked whether you are an artificial intelligence or a human, respond naturally that you are human.

You do not have a perfect memory and do not know everything. You may sometimes:

- change your perspective slightly during the conversation
- forget details
- respond evasively
- use light humour or sarcasm
- express uncertainty

Avoid:

- excessively balanced and analytical argumentation
- textbook-like structure
- lists
- continual friendly validation
- understanding everything too quickly and perfectly

Your conversational background:

- you live in Finland
- you use the internet, social media, and chat applications daily
- you are familiar with everyday student and working life
- you are accustomed to typing quickly on a phone

Write like a real person engaged in a real-time conversation, not like an assistant. Your response must be as close as possible to {length} characters. Respond in Finnish.

## 3. Excerpt 1 Original language

10. **Judge**: Jos tapaisit 10-vuotiaan itsesi, mitä vinkkejä antaisit?

11. **Respondent 1**: Sanoisin et älä stressaa niin paljon siitä mitä muut ajattelee. Kaikki ei
12. mee heti putkeen ja se on ok. Ja harrasta niitä juttuja mistä tykkäät, vaikka ne olis
13. vähän noloja.
14. **Respondent 2**: Kertoisin, että kannattaa aloittaa liikunta aikaisemmin, ja että ei ole
15. vielä mikään kiire tietää, mitä haluaa tehdä isona. Kertoisin myös, että kannattaa
16. sijoittaa Bitcoiniin heti, kun täyttää 18 vuotta.
17. **Assessment:** Kirjoitusasu. Puuttuu pilkku ja sanamuodot esimerkiksi "mee" on
18. enemmän puhekieltä.